\documentclass{article}

\usepackage{arxiv}

\usepackage[utf8]{inputenc} 
\usepackage[T1]{fontenc}    
\usepackage{hyperref}       
\usepackage{url}            
\usepackage{booktabs}       
\usepackage{amsfonts}       
\usepackage{nicefrac}       
\usepackage{microtype}      
\usepackage{lipsum}
\usepackage{moreverb,url}
\usepackage{graphicx}
\usepackage{amssymb,amsmath}
\usepackage{subcaption}
\usepackage{comment}
\usepackage{xcolor}
\usepackage{comment}

\title{Graph-Partitioning-Based Diffusion Convolutional Recurrent Neural Network for Large-Scale Traffic Forecasting}

\author{
  Tanwi Mallick \\
  Mathematics and Computer Science Division\\
  Argonne National Laboratory, Lemont, IL \\
  \texttt{tmallick@anl.gov} \\
   \And
 Prasanna Balaprakash \\
  Mathematics and Computer Science Division \& 
  Argonne Leadership Computing Facility \\
  Argonne National Laboratory, Lemont, IL \\
  \texttt{pbalapra@anl.gov} \\
  \And
 Eric Rask\\
  Energy Systems Division\\
  Argonne National Laboratory, Lemont, IL \\
  \texttt{erask@anl.gov} \\
  \And
  Jane Macfarlane\\
  Sustainable Energy Systems Group \\
  Lawrence Berkeley National Laboratory, Berkeley, CA \\
  \texttt{jfmacfarlane@lbl.gov} \\
}

\begin{document}
\maketitle

\begin{abstract}
Traffic forecasting approaches are critical to developing adaptive strategies for mobility. Traffic patterns have complex spatial and temporal dependencies that make accurate forecasting on large highway networks a challenging task. Recently, diffusion convolutional recurrent neural networks (DCRNNs) have achieved state-of-the-art results in traffic forecasting by capturing the spatiotemporal dynamics of the traffic. Despite the promising results, however, applying DCRNNs for large highway networks still remains elusive because of computational and memory bottlenecks. We present an approach for implementing a DCRNN for a large highway network that overcomes these limitations. Our approach uses a graph-partitioning method to decompose a large highway network into smaller networks and trains them independently. We demonstrate the efficacy of the graph-partitioning-based DCRNN approach to model the traffic on a large California highway network with 11,160 sensor locations. We develop an overlapping nodes approach for the graph-partitioning-based DCRNN to  include sensor locations from partitions that are geographically close to a given partition. Furthermore, we demonstrate that the DCRNN model can be used to forecast the speed and flow simultaneously and that the forecasted values preserve fundamental traffic flow dynamics. Our approach to developing DCRNN models that represent large highway networks can be a potential core capability in advanced highway traffic monitoring systems, where a trained DCRNN model forecasting traffic at all sensor locations can be used to adjust traffic management strategies proactively based on anticipated future conditions.
\end{abstract}


\section{Introduction}
In the United States alone, traffic congestion accounts for billions of dollars of economic loss due to productivity loss from additional travel time and additional inefficiencies and energy required for vehicle operation. To address these issues, intelligent transportation system (ITS) strategies~\cite{bishop2005intelligent}  seek to better manage and mitigate congestion and other traffic-related issues via a range of data-informed strategies and highway traffic monitoring systems. Near-term traffic forecasting is a foundational component of these strategies. Accurate forecasting across a range of normal, elevated, and extreme levels of congestion is critical for improved traffic control, routing optimization, and identification of novel approaches for handling emerging patterns of congestion \cite{teklu2007genetic, tang2005traffic}. Furthermore, predictions and models from machine learning methods can be used to delve more deeply into the dynamics of a particular transportation network in order to identify additional areas of improvement beyond those enabled by improved prediction and control \cite{fadlullah2017state, abdulhai2003reinforcement, lv2014traffic}. Forecasting methodologies are also expected to enable new forms of ITS strategies as they become integrated into larger optimization and control approaches and highway traffic monitoring systems \cite{pang1999adaptive, decorla1997total}. For example, the benefits of highly dynamic route guidance and alternative transit mode pricing in real time would be greatly aided by improved traffic forecasting. 

Traffic forecasting is a challenging  problem.  Key traffic metrics, such as flow and speed, exhibit complex spatial and temporal correlations that are difficult to model with classical forecasting approaches \cite{williams2003modeling, chan2012neural,karlaftis2011statistical, castro2009online}. From the spatial perspective, locations that are close geographically in the Euclidean sense (e.g., two locations located in opposite directions of the same highway) often do not exhibit a similar traffic pattern, whereas locations in the highway network that are relatively far apart (e.g., two locations separated by a mile in the same direction of the same highway) can show strong correlations. Many traditional predictive modeling approaches cannot handle these types of correlation. From the temporal perspective, because traffic conditions vary across different locations (e.g.,  diverse peak hour patterns, varying traffic flow and volume, highway capacity, incidents, and interdependencies), the time series data becomes nonlinear and nonstationary,  rendering many statistical time series modeling approaches ineffective.

Recently, machine learning, in particular, deep learning (DL), approaches have emerged as high-performing methods for traffic forecasting \cite{do2019survey}. Among these methods, the diffusion convolutional recurrent neural network (DCRNN) is a state-of-the-art method developed by Li et al.~\cite{li2017diffusion}for short-term traffic forecasting. DCRNN  models complex spatial dependencies using a diffusion process on a graph and temporal dependencies using a sequence to sequence recurrent neural network. 
Despite accurate forecasting results, however,  modeling large highway networks with DCRNN still remains challenging because of the computational and memory bottlenecks.  
In this paper, we focus on developing and applying DCRNN to a large highway network.
Our study is motivated by the fact that the highway network of a state such as California is $\approx$30 times larger than the size of the highway networks for which DCRNN was originally developed and demonstrated. 
The two main scaling challenges are that (1) the training data size for thousands of locations is too large to fit in a single compute node memory and (2) the time required for training a DCRNN on a large dataset can be prohibitive, rendering the method ineffective for large highway networks.
Distributed data-parallel and model-parallel training approaches  \cite{dean2012large} utilize multiple compute nodes to overcome these methods. 
In traditional computational science domains, a common approach for scaling is domain decomposition, wherein the problem is divided into a number of subproblems that are then distributed over different compute nodes. While domain decomposition approaches are not applicable in scaling typical DL training, such as image and text classification, they are well suited for the traffic forecasting problem with DCRNN because traffic flow in one region of the highway network does not affect another region when the regions are separated by a suitably large driving distance. To that end, we propose a graph-partitioning-based DCRNN approach that partitions a large highway network into subnetworks and trains a DCRNN for each subnetwork independently. In contrast to distributed data-parallel and model-parallel training approaches, multiple compute nodes are not a prerequisite for our method because the independent DCRNN models can be trained sequentially on a single compute node. Consequently, our approach is more amenable for implementation within traffic management centers (TMCs) 
without the need to access cloud computing resources. 
On the other hand, given such multinode cloud/computing access, our method can provide significant benefit with respect to overall model training time. Furthermore, we show that the short-term forecasting accuracy can be improved by partitioning the highway network, which results in smaller models that are easier to train. 
The main contributions of our work are as follows.

\begin{enumerate}
    \item We demonstrate the efficacy of the graph-partitioning-based DCRNN approach to model the traffic on a large California highway network with 11,160 sensor locations.
    \item We develop an overlapping nodes approach,  an improvement strategy for the graph-partitioning-based DCRNN that includes sensor locations from partitions that are geographically close to a given partition.
    \item We show that DCRNN can be extended for multioutput learning to forecast both flow and speed simultaneously, as opposed to the previous DCRNN implementation that forecast either speed or flow.
\end{enumerate}

\section{Related work}
Modeling the flow and speed patterns of traffic in a highway network has been studied for decades. Capturing the spatiotemporal dependencies of the highway network is a crucial task for traffic forecasting. 
The methods for short-term traffic forecasting are broadly classified into two  categories: knowledge-driven and data-driven approaches. In transportation and operational research, knowledge-driven methods  usually apply queuing theory \cite{cascetta2013transportation, romero2018queuing, lartey2014predicting, yang2014application} and Petri nets \cite{ricci2008petri} to simulate traffic behaviors. Usually, those approaches estimate the traffic flow of one intersection at a time. Traffic prediction for the full highway system of an entire state has not been attempted to date by using knowledge-driven approaches.

Data-driven approaches have received notable attention in recent years. The methods include statistical techniques such as autoregressive statistics for time series \cite{williams2003modeling} and Kalman filtering techniques \cite{kumar2017traffic}.  
These models are used mostly to forecast at a single sensor location and are based on a stationary assumption about the time series data. Therefore, they often fail to capture nonlinear temporal dependencies and cannot predict overall traffic in a large-scale network \cite{li2017diffusion}.  
Recently, machine learning methods on short-term traffic forecasting have emerged. More complex data modeling can be achieved by these models, such as support vector machines (SVMs) \cite{castro2009online, ahn2016highway} and artificial neural networks (ANNs) \cite{chan2012neural,karlaftis2011statistical}. However, SVMs are computationally expensive for large networks, and   ANNs cannot capture the spatial dependencies of the traffic network. Furthermore, the shallow architecture of ANNs make the network less efficient compared with a deep learning architecture. 

Recently, deep learning models such as deep belief networks~\cite{huang2014deep} and stacked autoencoders~\cite{lv2015traffic} have been used to capture effective features for short-term traffic forecasting. Recurrent neural networks (RNNs) and their variants, long short-term memory (LSTM) networks \cite{ma2015long} and gated recurrent units (GRUs) \cite{fu2016using}, show effective forecasting  \cite{ cui2018deep, yu2017deep} because of their ability to capture the temporal dependencies. However, spatial dynamics are often missed my RNN-based methods.
To capture the  spatial dynamics, researchers have used convolutional neural networks (CNNs).  Ma et al. \cite{ma2017learning} proposed an image-based traffic speed prediction method using CNNs, whereas Yu et al. \cite{yu2017spatiotemporal} proposed spatiotemporal recurrent convolutional networks for short-term traffic forecasting. 
Spatial dynamics have been captured by deep CNNs, and temporal dynamics have been learned by LSTM networks. In both, the highway network was represented as an image, and the link was colored by speed. The model was tested on only 278 links of the Beijing transportation network. 
Zhang et al. \cite{zhang2016dnn, zhang2017deep} also represented the flow of crowds in a traffic network using grid-based Euclidean space.  The temporal closeness, period, and trend of the traffic were modeled by using a residual neural network framework. The researchers evaluated the model on Beijing and  New York City crowd flows. They used two datasets:  trajectories of taxicab GPS data of four time intervals and  trajectories of NYC bikes during one interval of time. Trip data included trip duration,  starting and ending sensor IDs, and start and end times. 
The key limitation of these approaches is that they do not capture  non-Euclidean spatial connectivity. Du et al. \cite{du2018hybrid} proposed a model with one-dimensional CNNs  and GRUs with the attention mechanism used to forecast traffic flow on UK traffic data. The contribution of this method is multimodal learning by multiple feature fusion (e.g. flow, speed, events, weather)on time series data of one year. This dataset was composed of 34,876 15-minute intervals. However, it was limited to a narrow spatial dimension. 
 
Recently, CNNs have been generalized from a 2D grid-based convolution to a graph-based convolution in non-Euclidean space. Yu et al.\cite{yu2017spatio} modeled the sensor network as an undirected graph and proposed a deep learning framework, called a spatiotemporal graph convolutional network, for speed forecasting. They applied graph convolution and gated temporal convolution through spatiotemporal convolutional blocks. The experiments were done on two datasets: BJER4,  collected by the Beijing Municipal Traffic Commission, and  PeMSD7, collected by the California Department of Transportation. The maximum size of their dataset was 1,026 sensors of California District 7. However, these spectral-based convolution methods require the graph to be undirected. Hence, moving from a spectral-based to a vertex-based method, Atwood and Towsley \cite{atwood2016diffusion} first proposed convolution as a diffusion process across the node of the graph. Later, Hechtlinger et al. \cite{hechtlinger2017generalization} developed convolution to graphs by convolving every node and its closest neighbors selected by a random walk. However, none of these methods could capture the temporal dependencies.

Li et al. \cite{li2017diffusion}  used the DCRNN method to forecast performances for 15, 30, and 60 minutes on two datasets: a Los Angeles dataset with 207 locations collected over 4 months and a Bay Area dataset with 325 locations collected over 6 months. 
Their  results showed improvement on the state-of-the-art baseline methods such as historical average \cite{williams2003modeling}, an autoregressive integrated moving average model with a Kalman filter \cite{xu2017real}, a vector autoregressive model \cite{hamilton1995time}, a linear support vector regression, a feed-forward neural network \cite{raeesi2014traffic}, and an encoder-decoder framework using LSTM \cite{sutskever2014sequence}.

Our approach differs from these works in several respects. None have addressed a large set of sensor locations, whereas  we use 11,160 sensor locations that cover the major part of the California highway system. Moreover, our  graph-partitioning-based approach for large-scale traffic forecasting, using overlapping nodes, and multioutput multitask forecasting, using DCRNN, significantly extends the initial DCRNN idea to real-world solutions.

\section{Methodology}
In this section, we describe the DCRNN approach for traffic modeling, followed by graph partitioning for DCRNN, the overlapping node method, and multioutput learning.

\subsection{Diffusion convolutional recurrent neural network (DCRNN)}
\label{sec_dcrnn}

Formally, the problem of short-term traffic forecasting can be modeled as a spatial-temporal time series forecast defined on a  
weighted directed graph $G = (V, \varepsilon, A)$, where  $V$ is a set of $N$ nodes that represent sensor locations, $\epsilon$ is the set of edges connecting the sensor locations, and $A\in R^{N\times N}$ is the weighted adjacency matrix that represents the connectivity between the nodes in terms of highway network distance. Given the graph $G$ and the time series data $X_{t-T'+1}$ to $X_t$, the goal of the traffic forecasting problem is to learn a function $\text{h(.)}$  that maps historical data of $T'$ time steps at given time $t$ to future $T$ time steps.

\begin{equation*}
X_{t-T'+1}, ...,X_t; G \xrightarrow{\text{h(.)}} X_{t+1},... , X_{t+T}
\end{equation*}

In DCRNN, the temporal dependency of the historical data has been captured by the encoder-decoder architecture \cite{cho2014learning, sutskever2014sequence} of a recurrent neural network. The encoder steps through the input historical time series data and encodes the entire sequence into a fixed-length vector. The decoder predicts the output of the next $T$ time steps while reading from the vector. GRU \cite{cho2014learning} is used to design the encoder-decoder architecture. Inside the encoder-decoder architecture of RNN, a diffusion convolution operation is used to capture the spatial dependencies. The diffusion process \cite{teng2016scalable} can be described by a random walk on $G$.
The traffic flow from one node to the neighbor nodes can be represented as a weighted combination of infinite random walks on the graph. The matrix multiplications of RNN is replaced with the diffusion convolution operation to make the DCRNN cell, which is defined as
\begin{equation*}
\begin{array}{lcl}
r^t & = &  \sigma (W_{r\bigstar G} [X_{t}, h_{t-1}] + b_r) \\
u^t & = &  \sigma (W_{u\bigstar G} [X_{t}, h_{t-1}] + b_u) \\
c^t & = &  \tanh (W_{c\bigstar G} [X_{t} (r_t \odot h_{t-1}] + b_c) \\
h_t & = & u_t \odot h_{t-1} + (1 - u_t) \odot c_t ,
\end{array}
\end{equation*}

where $X_{t}$ and $h_t$ denote the input and final state
at time $t$, respectively; $r_t$, $u_t$, and $c_t$ are the reset gate, update gate, and cell state at time $t$;, respectively; $\bigstar G$ denotes the diffusion convolution operation; and $W_r, W_u,$ and $W_c$  are parameters for the corresponding filters. The diffusion convolution 
operation ($\bigstar G$) over the input graph signal $X$ is defined as

\begin{equation*}
 W_{\bigstar G} X = \sum_{d=0}^{K-1} (W_O (D_O^{-1}A)^d + W_{I}(D_I^{-1}A)^d) X,
\end{equation*}

where $K$ is a maximum diffusion steps;  $D_O^{-1}A$ and $D_I^{-1}A$ are transition matrices of the diffusion process and the reverse one, respectively; $D_O$ and $D_I$ are the in-degree and out-degree diagonal matrices, respectively, and $W_O$ and $W_I$ are the learnable filters for the bidirectional diffusion process. The in-degree and out-degree diagonal matrices provide the capability to capture the effect of the upstream as well as the downstream traffic.

During the training phase, the historical time series data and the graph are fed into the encoder, and the final stage of the encoder is used to initialize the decoder. 
The decoder predicts the output of the next $T$ time steps, and the layers of DCRNN are trained by using backpropagation through time. During the test, the ground truth observations are replaced by previously predicted output. The discrepancy between the input distributions of training and testing can cause performance degradation. In order to resolve this issue, scheduled sampling \cite{bengio2015scheduled} has been used, where the model is  fed a ground truth observation with a probability of $\epsilon_i$ or the prediction by the model with a probability of $1 - \epsilon_i$ at the $i$th iteration. The model is trained with a mean absolute error (MAE) loss function, defined as 

\begin{equation*}
    \text{MAE} = \frac{1}{s}\sum_{i=1}^s | y_i - \hat{y}_i |, 
\end{equation*}

where $y_i$ is the observed value,  $\hat{y}_i$ corresponds to the forecasted values for the $i$th training data, and $s$ denotes the number of samples. We refer the reader to Li et al. \cite{li2017diffusion} for a detailed exposition on DCRNN.

\subsection{Graph-partitioning-based DCRNN}
For the graph-partitioning-based DCRNN approach, we partition the the large graph $G$ into $k$ subgraphs such as $G = \{{G}_{0}, \ldots, {G}_{k-1}\} = [\{V_0, \varepsilon_0\}, \{V_2, \varepsilon_2\}, \ldots, \{V_{k-1}, \varepsilon_{k-1}\}]$,  where   each $\varepsilon_i$    consists   only of   the   edges   between   nodes   in $V_i$.  For  each  subgraph  we  have a set of historical time series data $X = \{X_{0}, \ldots, X_{k-1}\}$, and we  calculate  the  adjacency  matrix $A = [A_0, \ldots, A_k]$. Now, each subgraphs can be trained independently.

For selecting the graph partitions, various graph clustering and community detection methods \cite{liu2015empirical} have been developed, such as multilevel $k$-way graph partitioning  \cite{karypis1998multilevelk},  multilevel recursive bisection  \cite{karypis1998fast}, spectral clustering \cite{ng2002spectral}, and Louvain \cite{blondel2008fast}. 
We compared the different clustering methods and found that  multilevel $k$-way partitioning has several advantages over all other methods (See Supporting experiments, Impact of graph partitioning methods for the comparison). 
This method can partition a million-node graph in a few seconds with tightly connected clusters.  It takes the adjacency matrix as input and divides it into multiple partitions in three phases: (1) coarsening phase --  the graph is coarsened down to a smaller graph with fewer vertices and  a set of vertices is collapsed iteratively to form a single vertex; (2) initial partitioning phase -- the smaller coarsened graph is partitioned by using multilevel $k$-way partitioning \cite{karypis1998multilevelk}, where $k$ is the number of partitions; and (3) uncoarsening phase --  the partitions are projected back to the original graph by backtracking through the coarsened graph. In order to get tightly connected partitions, nodes are swapped between partitions by using the Kernighan-Lin algorithm \cite{hendrickson1995multi}.

\begin{figure}[t]
  \centering
    \includegraphics[width=0.9\linewidth]{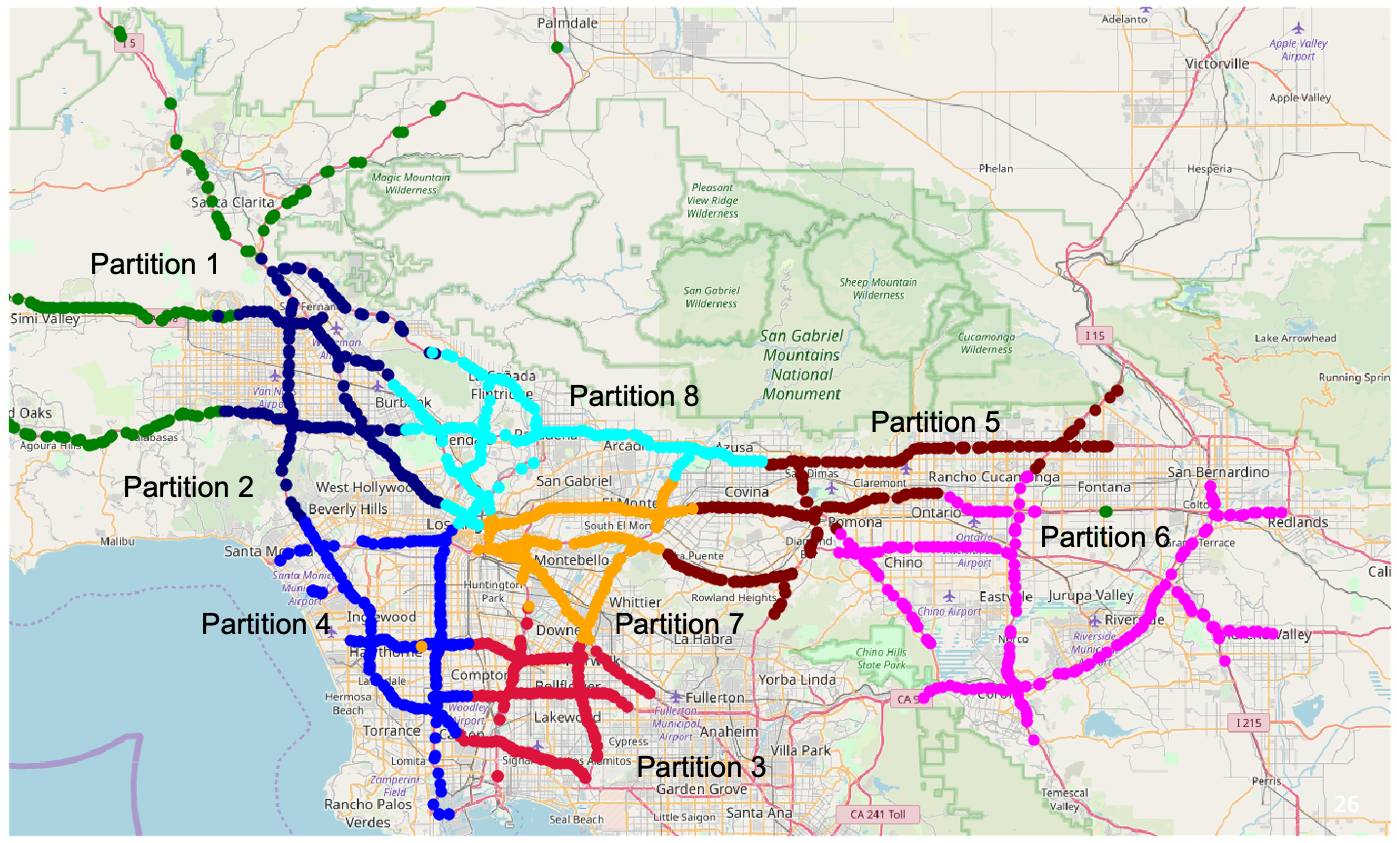}
    \caption{Result of $k$-way partitioning on Los Angles region: 2,036 sensor locations are partitioned into 8 parts, with each part rendered in different colors}
    \label{fig_LA_partition}
\end{figure}

An example of applying the $k$-way partitioning algorithm on the Los Angles region (2,036 sensors) is shown in  Figure \ref{fig_LA_partition}. Here, we partitioned the graph with 2,036 nodes into 8 parts by using the adjacency matrix as input to the partitioning algorithm and rendered the partitions in different colors. Each partition has approximately 254 nodes.

\subsection{Overlapping nodes}
\label{sec_overlap}

In Figure \ref{fig_LA_partition}, we  observe that nodes at the boundary of any partition lose their neighboring correlated nodes to nearby partitions despite the shorter driving distance. This issue will become critical and affect the prediction accuracy of DCRNN when the number of partitions increases.
To address this issue, we develop an overlapping nodes approach wherein, for each partition, we find and include spatially correlated nodes from other partitions. Consequently, the nodes that are near the boundary of the partition will appear in more than one partition.

\begin{figure}[t]
  \centering
    \includegraphics[width=0.9\linewidth]{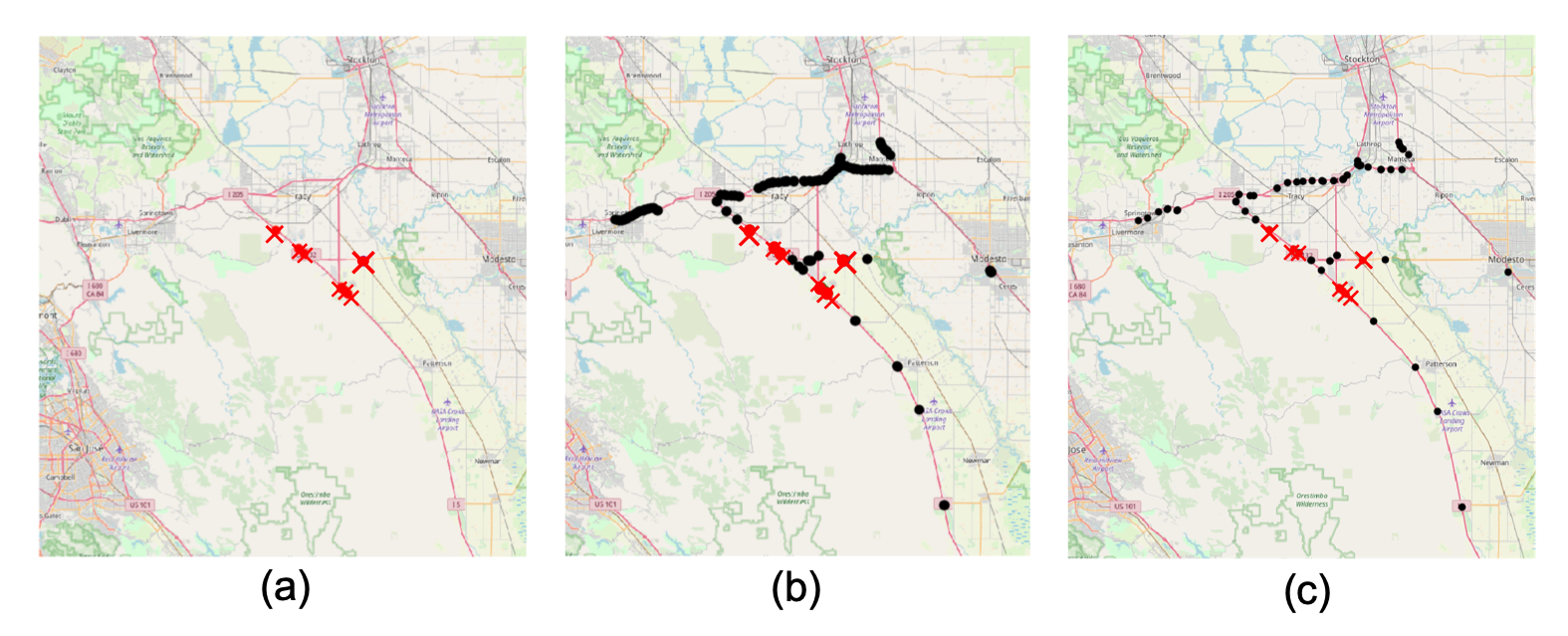}
    \caption{Example after adding overlapping nodes: (a) partition without overlapping nodes, (b)  partition after adding overlapping nodes, (c) partition after downsampling the overlapping nodes with distance threshold of 1 mile}
    \label{fig_sparse_example}
\end{figure}

A naive approach for finding the correlated nodes consists of computing nearest neighbors for each node in the partition based on the driving distance and excluding the nodes already included in the partition. The disadvantage of this approach is that it can include, for a given node, several spatially correlated nodes that are close to each other. This can lead to an increase in the number of nodes per partition and, consequently, higher training time and memory requirements. This issue is illustrated on a partition of the San Joaquin area, District 10 in Figure \ref{fig_sparse_example}, where the subfigure (a) shows a small partition with 11 nodes rendered in red cross symbols and the subfigure (b) shows same partition after adding the 162 overlapping nodes rendered in black dot symbol.  After adding the overlapping nodes, the total number of nodes become 15 times more than the original partition.
Therefore, we downsample the spatially correlated nodes from other partitions as follows: Given two spatially correlated overlapping nodes from a different partition, we select only one and remove the other if they are within $D'$ driving distance miles, where $D'$ is a parameter.

The subfigure (c) of Figure \ref{fig_sparse_example} shows the results after downsampling with a distance threshold of 1 mile. The total number of nodes becomes 40 after after downsampling.

\subsection{Multioutput forecasting with a single model}
\label{sec_multioutput}

Previously, DCRNN was used to forecast speed based on historical speed data. We customized the input and output layers of the DCRNN for multioutput forecasting such that a single DCRNN model can be trained and used for forecasting speed and flow simultaneously. The three key modifications were needed for multioutput forecasting. 
(1) Normalization of speed and flow: 
To bring speed and flow to the same scale, normalization was  done separately on the two features using the standard scalar transformation. The normalized values of speed are given by $z_{sp} = \frac{x_{sp} - \mu_{sp}}{\sigma_{sp}}$, where $\mu_{sp}$ is the mean and $\sigma_{sp}$ is the standard deviation of the speed values $x_{sp}$. The same method was applied for normalizing the flow values ($z_{fl} = \frac{x_{fl} - \mu_{fl}}{\sigma_{fl}}$, where $\mu_{fl}$ and $\sigma_{fl}$ are the standard deviation of the flow values $x_{fl}$). We applied an inverse transformation to the normalized speed and flow forecasting values to transform them to the original scale (for computing error on the test data);
(2) Multiple output layers in the DCRNN: In the previous study of DCRNN, the diffusion convolution layer learns P-dimensional input, such as speed and flow, and predicts Q-dimensional output, such as speed and flow. 
(3) Loss function: For multioutput training, we use a loss function of the form 
\begin{equation*}
\text{MAE}_{\text{multi}} = \text{MAE}_{sp} + \text{MAE}_{fl}  
= \frac{1}{s}\sum_{i=1}^s | y_{sp_{i}} - \hat{y}_{sp_{i}} | + \frac{1}{s}\sum_{i=1}^s | y_{fl_{i}} - \hat{y}_{fl_{i}}|, 
\end{equation*}
where $y_{sp_{i}}$ and $y_{fl_{i}}$ are observed speed and flow values, respectively;  $\hat{y}_{sp_{i}}$ and $\hat{y}_{fl_{i}}$ are corresponding forecast values, respectively, for the $i$th training data; and $s$ is the total number of training points.

\section{Experimental Results}

First, we describe the dataset of the California highway network used for our experiments. Next, we evaluate the efficacy of the DCRNN model with respect to the number of partitions and analyze the model errors using a sensitivity analysis approach. Then, we study the impact of overlapping nodes and of hyperparameter search on the DCRNN accuracy. We conclude the section with a demonstration of multioutput forecasting using a graph-partitioning-based DCRNN.

\subsection{Dataset: California highway network}
\label{sec_data_cal}
We evaluated our approach on the California highway network. We used data from PeMS \cite{pems}, which provides access to real-time and historical performance data from over 39,000 individual sensors. The individual sensors placed on the different highways are aggregated across several lanes and are fed into vehicle detector stations. 
Vehicle detector stations include a variety of sensors such as inductive loops, radar, and magnetometers. The sensors may be located on high-occupancy vehicle (HOV) lanes, mainlines, on-ramps, and off-ramps. The dataset covers 9 districts of California---D3 (North Central) with 1,212 stations,  D4  (Bay Area) with  3,880 stations, D5 (Central coast) with 382 stations, D6 (South Central) with 624 stations, D7 (Los Angeles) with 4,864 stations, D8 (San Bernardino) with 2,115 stations, D10 (Central) with 1,195 stations, D11 (San Diego) with 1,502 stations, and D12 (Orange County) with 2,539 stations---giving a total of 18,313 stations listed by location.  Detectors capture samples every 30 seconds. PeMS then aggregates that data to the granularity of 5 minutes, an hour, and a day. The data include timestamp, station ID, district, freeway, direction of travel, total flow, and average speed (mph). 

PeMS does not list the latitude and longitude for the stations IDs, which are essential for defining the connectivity matrix used by DCRNN. Instead, the latitude and longitude are associated with postmile markers of every freeway given the direction. We 
found the latitude and longitude for sensor IDs by matching the absolute postmile markers of every freeway. Linear interpolation was used to find the exact latitude and longitude if the absolute postmile markers did not match exactly. 

The official PeMS website shows that 69.59\% of the $\approx$18K stations are in good working condition. The remaining 30.41\% do not capture time series data throughout the year and therefore were excluded from our dataset. Our final dataset has speed and flow of 11,160 stations from January 1, 2018, to December 31, 2018, with a granularity  of every 5 minutes. 

We observed that flow and speed values are missing for multiple time periods in the time series data. The percentage of missing data is  small---for speed 0.06\% (698,162 out of 1,173,139,200 data points) and for flow 0.04\% (504,688 out of 1,173,139,200 data points). We explored three different imputation techniques---temporal mean, temporal median, and linear interpolation---and did not find significant difference because the percentage of missing data is less than 1\%. The details of the experiments and results are reported in the Supporting experiments, Missing data imputation. Therefore, we replaced the missing data by the temporal mean.

\subsection{Experimental setup}
We represented the highway network of 11,160 sensors as a
weighted directed graph. From the one-year data, we used the first 70\% of the data (36 weeks approx.) for training and the next 10\% (5 weeks approx.) and 20\% (10 weeks approx.) of the data for validation and testing, respectively.
We experimented with different training data sizes by selecting 2,  4,  12,  20, and 36 weeks of data and found that using 36 weeks results in better accuracy.
Given 60 minutes of time series data on the nodes in the graph, we forecast for the next 60 minutes. We prepared the dataset in a way to look back ($T'$) for 60 minutes or 12 time steps (granularity of the data is 5 minutes) to predict ($T$) the next 60 minutes or 12 time steps. The $T'$ window slides by 5 minutes or 1 time step and repeats until all the available data is consumed. The forecasting performance of the models was evaluated on the test data by using MAE. 

We used the Open Source Routing Machine (OSRM) \cite{osrm} to compute  the highway network distance between the nodes, which is required by the adjacency matrix for DCRNN.  Given the latitude and longitude of two nodes, a locally running OSRM gives the shortest driving distance between them using OpenStreetMap \cite{osm}. To speed up the  computation, first we found 30 nearest neighbors for each node using the Euclidean distance and then limited the OSRM queries only to the nearest neighbors. As in the original DCRNN work, we built the adjacency matrix using a thresholded Gaussian kernel \cite{shuman2012emerging}: $A_{ij} = \exp (- \frac{ dist(v_i, v_j)^2}{\sigma^2})\;\; if \;\; dist(v_i, v_j)^2 \leq thresh,$ otherwise $0$, where $A_{ij}$ represents the edge weight between node $v_i$ and node $v_j$; $dist(v_i, v_j )$ denotes the highway network distance from node $v_i$ to node $v_j$; $\sigma$ is the standard deviation of distances; and $thresh$ is the threshold, which introduces the sparsity in the adjacency matrix. 

For the experimental evaluation, a GPU-based cluster at the Argonne Leadership Computing Facility was used. It has 126 compute nodes, where each node consists of two 2.4 GHz Intel Haswell E5-2620 v3 processors (6 cores per CPU, 12 cores total), one NVIDIA Tesla K80 
(two GPUs per node), 384 GB RAM per node, and 24 GB GPU RAM per node (12 GB per GPU). We used Python 3.6.0, TensorFlow 1.3.1, and Metis 5.1.0.  DCRNN code of \cite{li2017diffusion}, available on GitHub \cite{li2018dcrnn_traffic}, was customized for this implementation. Given $k$ partitions of the highway network, we trained partition-specific DCRNNs simultaneously on GPU nodes. We used two MPI ranks per node, where each rank ran a partition-specific DCRNN using one GPU. The input data for different partitions (time series and adjacency matrix of the graph) was prepared offline and loaded into the partition-specific DCRNN before the training started. We note that such simultaneous training is not required, in particular when a multi-GPU cluster is not available. The training on a single GPU would consist of running the training for each partition sequentially.

We used the default hyperparameter configuration values for the DCRNN:  batch size -- 64; filter type -- random walk,  used to model the stochastic nature of highway traffic; number of diffusion steps -- 2; RNN layers -- 2;  RNN units per layer -- 16;  threshold for gradient clipping -- 5; initial learning rate -- 0.01; and  learning rate decay --  0.1. We trained our model by minimizing MAE using the Adam optimizer \cite{kingma2014adam}.

\subsection{Number of partitions} \label{sec_partition_impact}

\begin{figure}[t]
  \centering
    \includegraphics[width=1\linewidth]{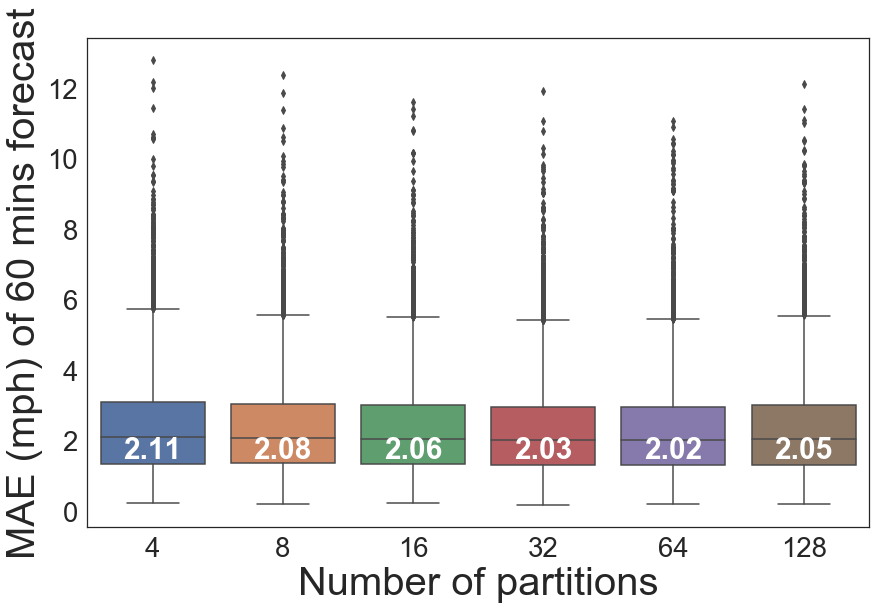}
    \caption{Distribution of MAE for different number of partitions}
    \label{fig_accuracy}
\end{figure}


Here, we demonstrate the efficacy of the graph-partitioning-based DCRNN.

The multilevel $k$-way graph-partitioning method from the Metis software package \cite{metis} was used to create 2, 4, 8, 16, 32, 64, and 128 partitions of the California highway network graph. The average number of nodes in each cases was 5,580, 2,790, 1395, 697, 348, 174, and 87, respectively. Partitions of size 1, the full network, and 2 are not presented because the training data was too large to fit in the memory of a single K80 GPU node. Given $k$ partitions, we used $k/2$ nodes (or $k$ GPUs) to run the partition-specific DCRNNs simultaneously.  The training time is defined as the maximum time taken by any partition-specific DCRNN training, excluding the data loading time.

Figure \ref{fig_accuracy} shows the distribution of MAE of all 11,160 nodes obtained by using box-and-whisker plots. 
From the results we can observe that medians, 75\% quantiles, and the maximum MAE values show a trend in which an increase in the number of partitions decreases the MAE.
From 4 to 64 partitions, the median of MAE decreases from 2.11 to 2.02. The increase in accuracy can be attributed to two factors: (1) the effectiveness of the $k$-way graph partitioning of Metis that separates sensor locations that were far apart with respect to driving distance and (2) the fact that the model training for each partition becomes relatively easier as the number of nodes in a given partition decreases.
For 128 partitions (with only 87 nodes per partition), the observed MAE values are higher than for 64 partitions. The reason is that the graph partition results in a significant number of spatially correlated nodes ending up in different partitions. This can be assumed as a tipping point for graph partitioning, which relates to the size and spread of the actual network.

\begin{figure}
  \centering
    \includegraphics[width=1\linewidth]{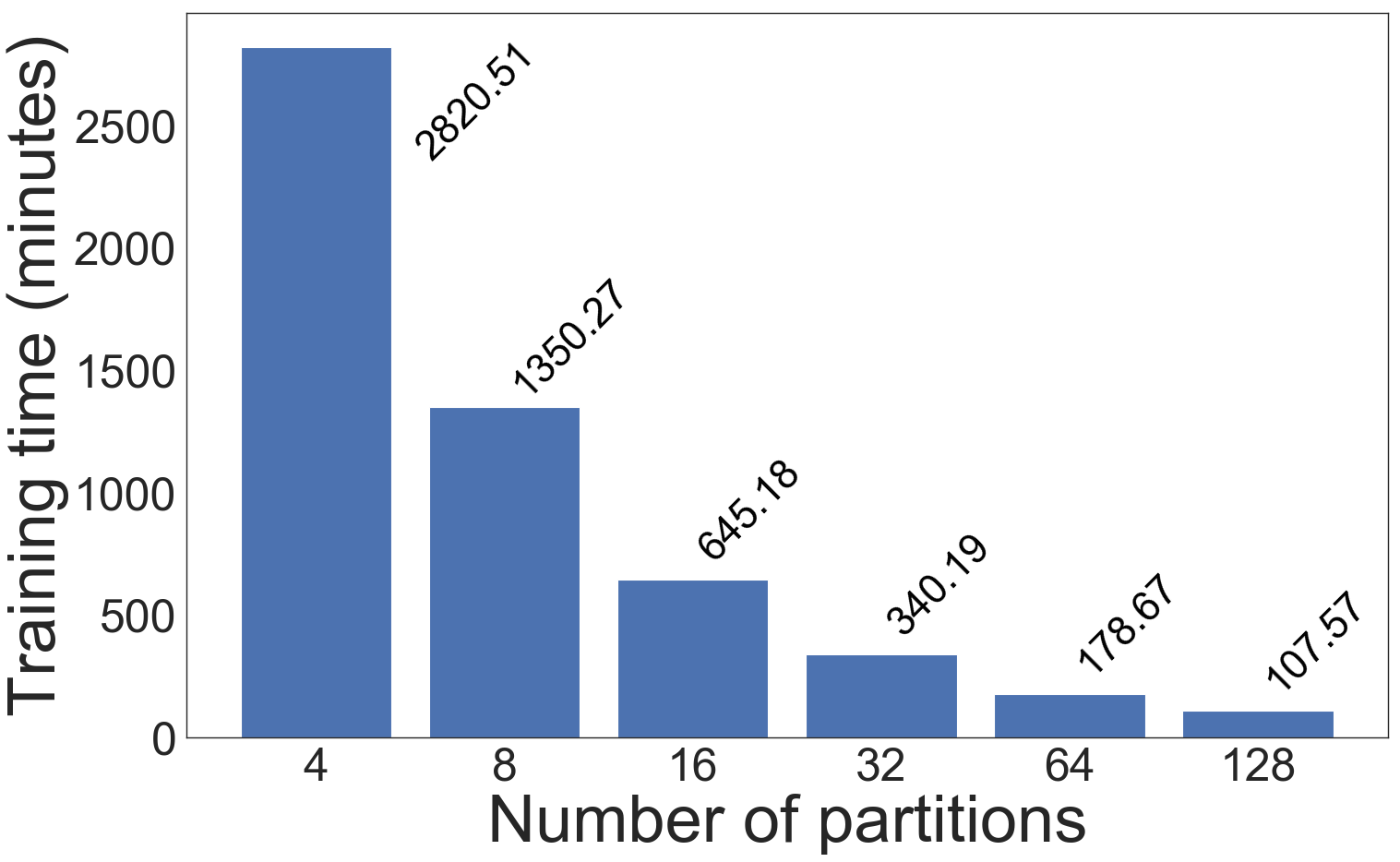}
    \caption{Training time for DCRNNs with different numbers of partitions}
    \label{fig_strong_scaling_time}
\end{figure}

Figure \ref{fig_strong_scaling_time} shows the training time required for different numbers of partitions. The time decreases significantly with an increase in the number of partitions. In addition, our approach reduces the training time from 2,820 minutes on 4 partitions (= 4 GPUs) to 178.67 minutes on 64 partitions (= 64 GPUs), resulting in a 15.78x speedup.  There is an almost linear speedup until 64 partitions, where doubling the number of partitions (and GPUs) results in $\approx$2X speedup. However, the speedup gains drop significantly with 128 nodes. At this point, with only 87 nodes per partition there is not enough workload for the GPU.

Since the best forecasting accuracy and speedup were obtained by using 64 partitions, we used it as a default number of partitions in the rest of the experiments. 

\subsection*{Error analysis} \label{sec_error_analysis}

\begin{figure}[!t]
  \centering
    \includegraphics[width=1\linewidth]{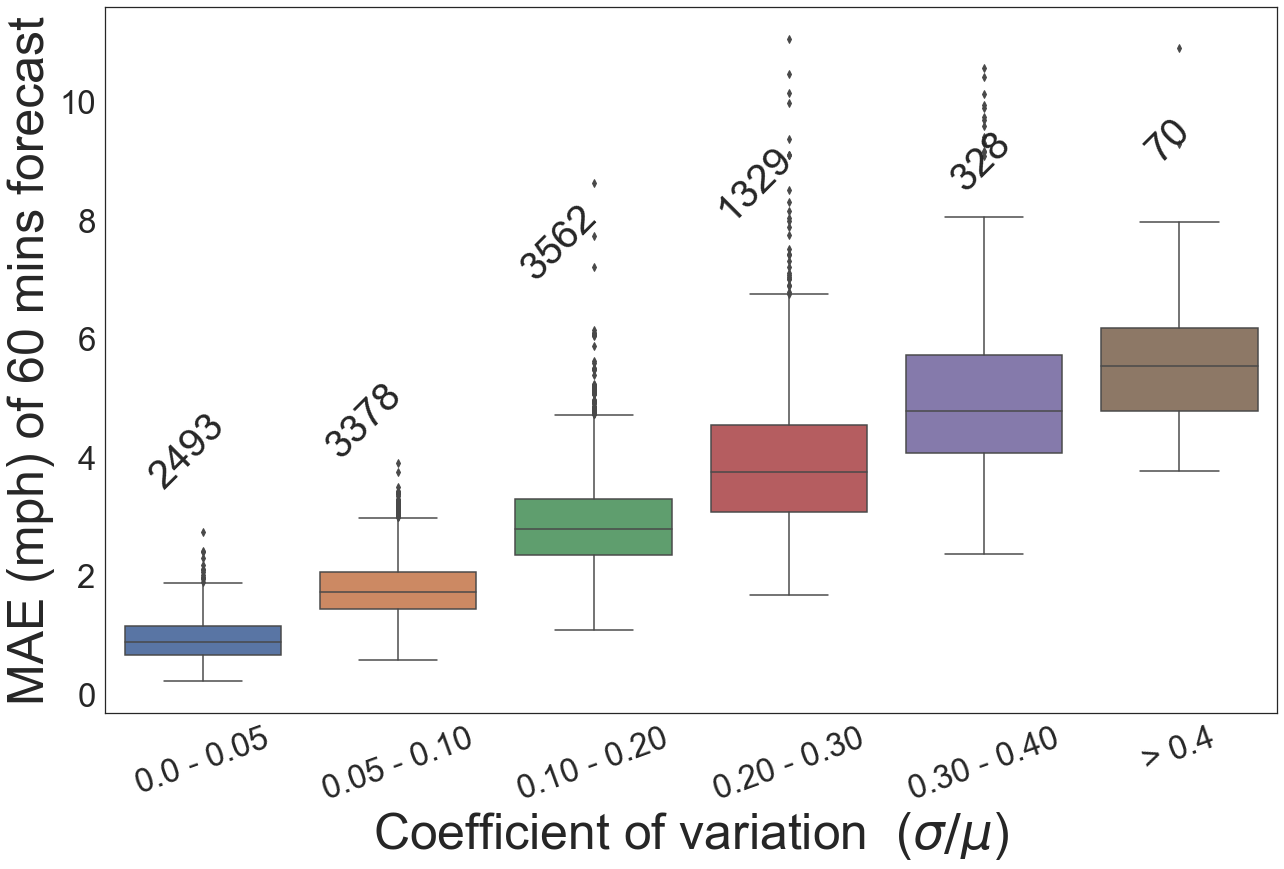}
    \caption{Impact of the coefficient of variation of the data on the forecasting accuracy. The  coefficient of variation of 11,160 sensors is binned into 6 categories. The number of nodes in each distribution is shown above each box.}
    \label{fig_variance}
\end{figure}

Figure \ref{fig_accuracy} shows several outliers (large  errors) in the MAE distribution. Here, using a decision tree method for sensitivity analysis, we investigate the factors that led to these large errors.

The factors that we studied are (1) sensor type (loop detector, magnetometers, etc.), (2) district where the sensor is located (Los Angeles, Bay Area, etc.), (3) lane type (mainline, HOV, etc.), and (4) traffic dynamics at a given sensor location measured by the coefficient of variation (standard deviation ($\sigma$)/ mean ($\mu$) of the time series data). The coefficient of variation becomes large (small) when the variations in the speed values are large (small). For each node, the values of the four factors are used as input (independent variables), and the corresponding forecasting error values (MAE) from 64 partition DCRNN are used as output (dependent variables). The MAE values are separated into 4 classes: class 0 if MAE less than 1, class 1 if MAE is between 1 and 3, class 2 if MAE is between 3 and 5, and class 3 if MAE is greater than 5. Consequently, for each node we have an input-output pair; the dataset comprises the pair from all the nodes.

Next, we trained a decision tree \cite{decisiontree} from the scikit-learn package, a supervised machine learning method, to model the error classes as a function of the four factors. We selected the decision tree because of the model interpretability, which can be used for analyzing the factors under study. We used 80\% of the data for training and 20\% for testing. To avoid overfitting, we set  the depth of the tree to 8; all other values were set to the default as in scikit-learn decision tree interface. The decision tree model obtained 82.11\% and 78.18\% accuracy on the training and testing sets, respectively. The high accuracy indicates that the error classes can be modeled and explained through the  four factors.

The trained decision tree model provides a normalized importance score (between 0 and 1) for each factor. The values for traffic dynamics, district, sensor type, and lane type were 0.87,  0.07,  0.04,  and 0.02, respectively. These results show that the traffic dynamics has the most significant impact on the forecasting error and that the other factors do not have impact. Figure \ref{fig_variance} shows the distribution of MAE values with respect to the coefficient of variation. We observe a clear trend in which the median (and other quantiles) of the MAE distribution increases linearly with respect to the  coefficient of variation.

To take a closer look at the sensors with high errors, we analyzed  class 3 (MAE > 5). This class had 449 sensors,  approx. 4\% of the total data (449 out of 11,160 sensors).
Of these 449 sensors, 244  were from  D7 (Los Angeles), 85  from  D4 (Bay Area), 42  from  D8 (San Bernardino), and 36   district D12 (Orange County). 
Given our forecasting horizon of 60 minutes, forecasting becomes difficult for the nodes with a high coefficient of variation.
For these nodes, the error decreases with a decrease in the forecasting horizon. For the 30- and 15-minute forecast horizons, only 125 and 36 sensors have MAE values that are greater 5 mph. From these results, we infer that for the nodes with high traffic dynamics, good forecasting accuracy can be achieved by reducing the forecasting horizon to 15 minutes.

\subsection{Impact of overlapping nodes}
Here, we discuss the impact of overlapping nodes on the forecasting accuracy of the graph-partitioning-based DCRNN. 

\begin{table*}[t]\begin{center}
\begin{tabular}{ll|r|r|r|r|r|r|r|}
\cline{3-9}
 &  & \multicolumn{1}{c|}{\begin{tabular}[c]{@{}c@{}}MAE \\ \textless 1\end{tabular}} & \multicolumn{1}{c|}{\begin{tabular}[c]{@{}c@{}}1 \textless{}= \\ MAE \\ \textless 3\end{tabular}} & \multicolumn{1}{c|}{\begin{tabular}[c]{@{}c@{}}3 \textless{}= \\ MAE \\ \textless 5\end{tabular}} & \multicolumn{1}{c|}{\begin{tabular}[c]{@{}c@{}}MAE \\ =\textgreater 5\end{tabular}} & \multicolumn{1}{c|}{\begin{tabular}[c]{@{}c@{}}Trainable \\ parameters\end{tabular}} & \multicolumn{1}{c|}{\begin{tabular}[c]{@{}c@{}}Training\\ time (m)\end{tabular}} & \multicolumn{1}{c|}{\begin{tabular}[c]{@{}c@{}}Forecast\\ time (m)\end{tabular}} \\ \hline
\multicolumn{1}{|l|}{1.} & \texttt{DCRNN\_64\_naive} & 1,716 & 6,729 & 2,266 & 449 & 14,608 & 178.67 & 4.38 \\ \hline
\multicolumn{1}{|l|}{2.} & \begin{tabular}[c]{@{}l@{}}\texttt{DCRNN\_64\_overlap} \\ \end{tabular} & 1,837 & 6,687 & 2,204 & 432 & 14,608 & 221.04 & 4.88 \\ \hline
\multicolumn{1}{|l|}{3.} & \begin{tabular}[c]{@{}l@{}}\texttt{DCRNN\_64\_naive\_hps} \\ \end{tabular} & 1,920 & 6,897 & 1,980 & 363 & 19,808 & 287.05 & 4.92 \\ \hline
\multicolumn{1}{|l|}{4.} & \begin{tabular}[c]{@{}l@{}}\texttt{DCRNN\_64\_overlap\_hps} \\ \end{tabular} & 1,897 & 6,940 & 1,972 & 351 & 38,048 & 461.57 & 5.83 \\ \hline
\end{tabular}
\caption{Results of graph-partitioning-based DCRNN with overlapping nodes and hyperparameter search}
\label{tab_HPS_results}
\end{center}
\end{table*}

\begin{figure}[h!]
\centering
    \includegraphics[width=1\linewidth]{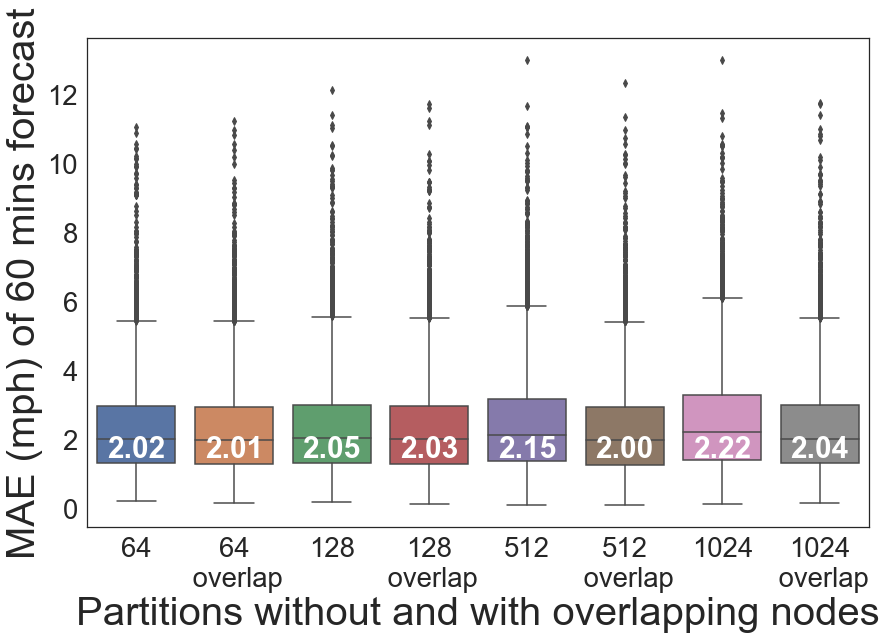}
    \caption{Impact of overlapping nodes in different partition scales}
  \label{fig_impact_overlap}
\end{figure}

Table \ref{tab_HPS_results} describes four different experiments.
The graph-partitioning-based DCRNN was trained on 64 partitions of the California highway network. We refer to this variant as\texttt{DCRNN\_64\_naive}. Training time took 178 minutes. 
After training, we forecast the speed for 60 minutes on the test data and calculated the MAE for each node. The results are summarized in the first row of Table \ref{tab_HPS_results} and the 64 partition results in Figure \ref{fig_impact_overlap}. We observe that the MAE values of 1,716, 6,729, 2,266, and 449 nodes are less than 1, between 1 and 3, between 3 and 5, and greater than 5, respectively. 

We also trained the graph-partitioning-based DCRNN on 64 partitions with overlapping nodes. 
We downsampled nodes with different distance threshold ($D'$) values: 0.5 mile, 1 mile, 1.5 miles, 2 miles, and 3 miles. The results showed no significant improvement beyond the 1 mile threshold; therefore, we used 1 mile as the distance threshold for our experiments. In a given partition, while calculating the MAE value for each node, we did not consider the overlapping nodes because they are included in a different partition, where their MAE values will be computed. We refer to this variant as \texttt{DCRNN\_64\_overlap}. The results are shown in  row 2 of Table \ref{tab_HPS_results}. We observe that \texttt{DCRNN\_64\_overlap} outperforms \texttt{DCRNN\_64\_naive}. With reference to the latter, the number of nodes with MAE values less than 1 has increased from 1,716 to 1,837; similarly, the number of nodes with MAE values between 1 and 3, 3 and 5, and greater than 5 decreased from 6,729 to 6,687, from 2,266 to 2,204, and from 449 to 432, respectively. We observe that the training time increased from 178.67 minutes to 221.04 minutes, which can be attributed to the increase in the number of nodes per partition.

Earlier, we noted that the forecasting accuracy decreased after increasing the number of partitions from 64 to 128 as a significant number of spatially correlated nodes ended up in different partitions. We hypothesize that including overlapping nodes can improve forecasting accuracy significantly for those partitions. Therefore, to analyze the impact of the overlapping nodes in different partition scales, we considered 512 and 1,024 partitions where each graph partition contains approximately 21 and 11 nodes, respectively. We added the overlapping nodes to each partition by downsampling them with a 1-mile distance threshold. We then trained and tested the graph-partitioning-based DCRNN model for 64, 128, 512, and 1,024 partitions with and without overlapping nodes. Figure \ref{fig_impact_overlap} shows the distribution of MAE values at each scale with and without overlapping nodes. The median of MAE decreases from 2.02 to 2.01 for 64 partitions, from 2.05 to 2.03 for 128 partitions, from 2.15 to 2.00 for 512 partitions, and from 2.22 to 2.04 for 1,024 partitions. These results show that the impact of overlapping nodes becomes more significant when the subgraph becomes very small and spatially correlated neighboring nodes belong to different partitions.

\subsection{Impact of hyperparameter tuning}
Here, we investigate the impact of hyperparameter search on the forecasting accuracy of the graph-partitioning-based DCRNN.

The hyperparameters that can impact the forecasting accuracy of the DCRNN include batch size, filter type (i.e., random walk, Laplacian), maximum diffusion steps, number of RNN layers, number of RNN units per layers, a threshold \texttt{max\_grad\_norm} to clip the gradient norm to avoid exploring gradient problem of RNN \cite{pascanu2013difficulty}, initial learning rate, and learning rate decay.


We used DeepHyper \cite{balaprakash2018deephyper}, a scalable hyperparameter search (HPS) package for neural networks, to search for high-performing hyperparameter values for 
\texttt{DCRNN\_64\_naive} and \texttt{DCRNN\_64\_overlap}. We used 5 months of data (from May 2018 to October 2018) from partition 1. We used 32 nodes with a 12-hour  wall-clock time as the stopping criterion. DeepHyper sampled 518 and 478 hyperparameter configurations for naive and overlapping approaches, respectively. The best hyperparameter configurations were selected from each and used to train the models and infer the forecasting accuracy.  We refer to these two variants as \texttt{DCRNN\_64\_naive\_hps} and \texttt{DCRNN\_64\_overlap\_hps}. The results are shown in rows 3 and 4 of Table \ref{tab_HPS_results}. We observe that \texttt{DCRNN\_64\_naive\_hps} outperforms \texttt{DCRNN\_64\_naive}, where hyperparameter tuning improved the accuracy of several nodes. The number of nodes with MAE values less than 1 and between 1 and 3  increased from 1,716 to 1,920 and from 6,729 to 6,897, respectively. The number of nodes with MAE values between 3 and 5 and greater than 5 decreased from 2,266 to 1,980 and from 449 to 363, respectively. 
We  also observe a similar trend with the comparison of \texttt{DCRNN\_64\_overlap} and \texttt{DCRNN\_64\_overlap\_hps}.
In particular, the number of nodes with MAE values between 3 and 5 and greater than 5 decrease from 2,204 to 1,972 and from 432 to 351, respectively. Moreover, hyperparameter tuning resulted in an increase in the number of trainable parameters, which led to an increase in training time  from 221.04 minutes to 461.57 minutes. This is because with overlapping nodes, DCRNN models already reach high accuracy; further improvements with hyperparameter search require a significant increase in the number of trainable parameters, which increases the training time. We did not notice any significant difference in the time required for forecasting by the different trained models on the test data. An exception is the \texttt{DCRNN\_64\_overlap\_hps} case, where the large number of additional trainable parameters increases the forecasting time by 1 minute (5.83 min).

Next, we tested whether  hyperparameter search on individual partitions can improve the accuracy further. We selected two partitions (partition 38 from the Bay Area with 179 sensors and partition 62 from South Central with 184 sensors) from 64 partitions and ran a hyperparameter search using the same setup.  The best hyperparameter configurations were selected for each partition and used to train and infer the forecasting accuracy on the same partition. We compared the results of hyperparameter tuning on individual partitions with results of \texttt{DCRNN\_64\_naive\_hps} on the same partitions. For partition 38,  the number of nodes with MAE values less than 1, between 1 and 3, and between 3 to 5  increased from 5 to 6, from 118 to 119, and from 38 to 41, respectively. The number of nodes with MAE values greater than 5 decreased from 18 to 13. Similarly, for partition 62,  the number of nodes with MAE values less than 1 and between 1 and 3 increased from 46 to 48 and from  121 to 123, respectively. The number of nodes with MAE values between 3 to 5  decreased from 13 to 9;  nodes with MAE values greater than 5 had no effect and remained at 4. The number of trainable parameters for partition 38 and 62 were 19,520 and 37,856, respectively. The results show that the hyperparameter search on individual partitions can improve the accuracy slightly but not significantly.

\subsection{Multioutput forecasting}

\begin{figure}[t]
    \includegraphics[width=1\linewidth]{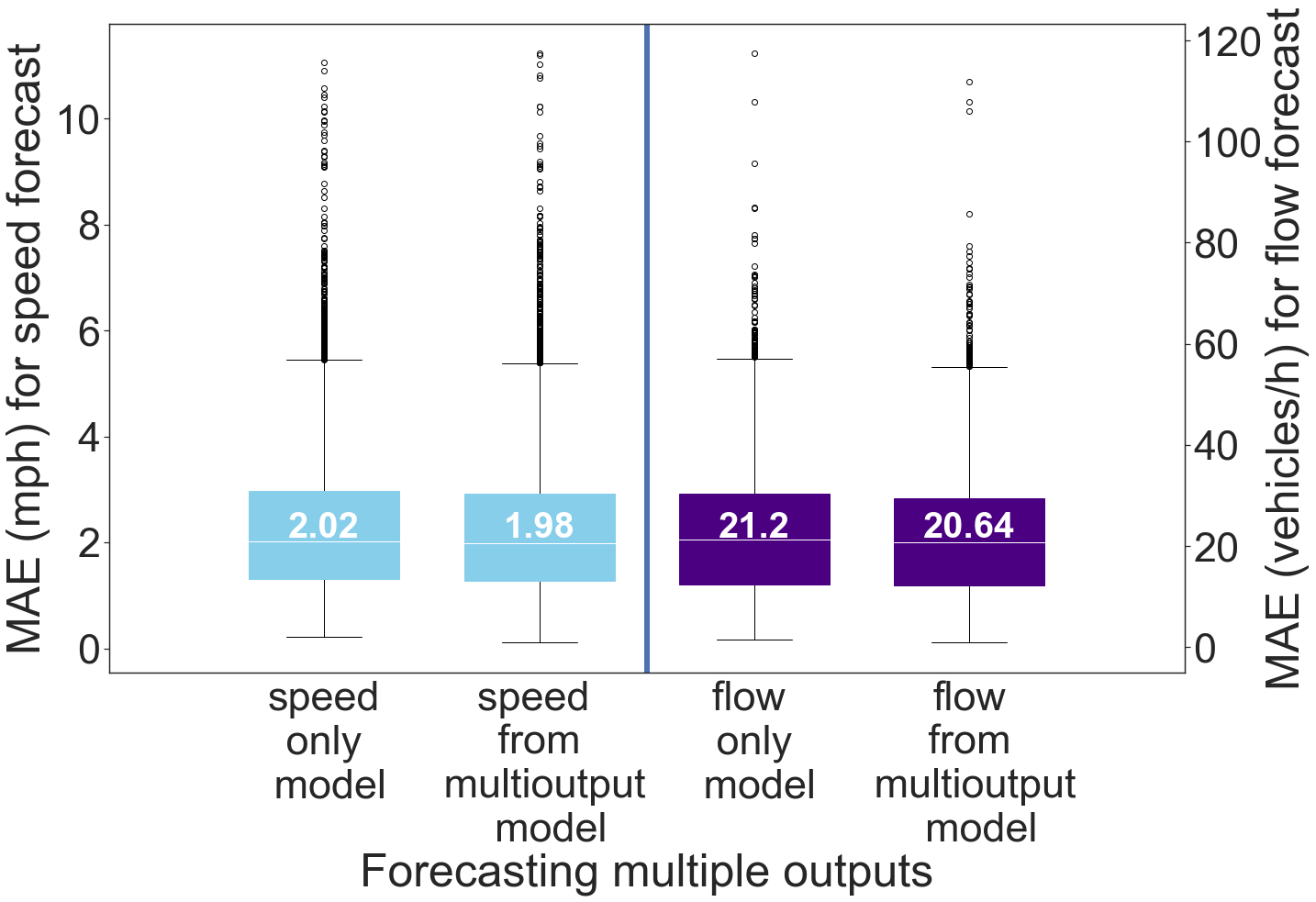}
    \caption{Box plot distribution of MAE for speed and flow forecasting. From left to right the box plots show the results of: speed forecasting from speed only model, speed forecasting from multioutput model, flow forecasting from flow only model, and flow forecasting from multioutput model}
  \label{fig_mult_output}
\end{figure}

We trained the graph-partitioning-based DCRNN model to forecast both speed and flow simultaneously and compared the results with those of models that predict either speed or flow. Figure \ref{fig_mult_output} shows the distribution of MAE values.
The median of MAE from the speed-only model (first box plot) is 2.02, which was reduced to 1.98 with the multioutput model (second box plot). Similarly, the median of MAE from the flow-only model (third box plot) is 21.20, which was reduced to 20.64 with the multioutput model (fourth box plot). 
We  found that the multioutput model obtains MAE values that are significantly lower  than those of the speed-only or flow-only models (a paired t-test $p-$values of $9.20 \times 10^{-4}$ for speed and $5.77 \times 10^{-5}$ for flow).
The superior performance of multioutput forecasting can be attributed to multitask learning \cite{sener2018multi}. 
The key advantage is that it leverages the commonalities across speed and flow learning tasks, which are related. This results in improved learning efficiency and consequently improved forecasting accuracy when compared with training the models separately.

\begin{figure*}[t]
     \centering
     \begin{subfigure}[b]{0.32\textwidth}
         \centering
         \includegraphics[width=\textwidth]{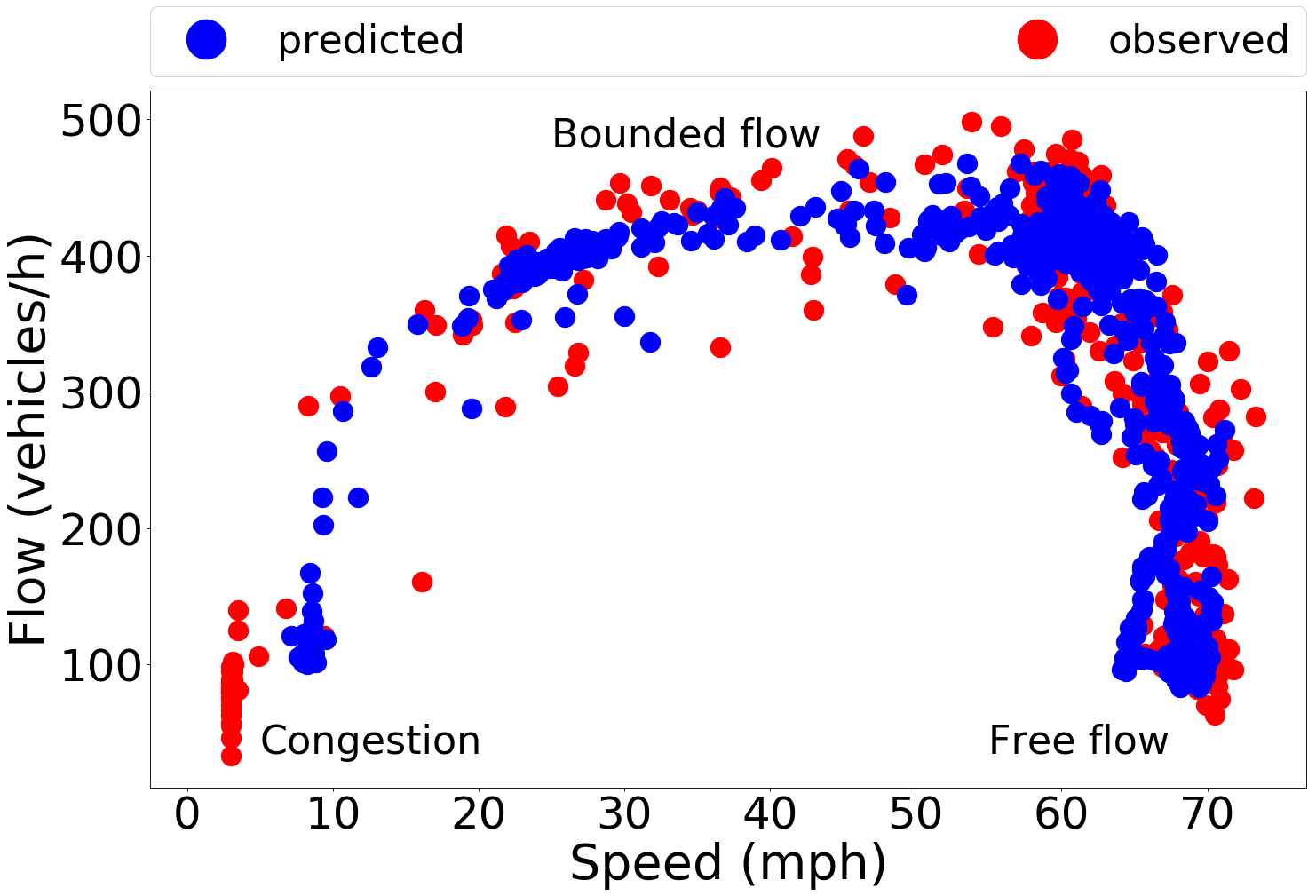}
         \caption{Closeness of the predicted flow and speed on sensor 717322 with fundamental traffic flow diagram}
         \label{fig_flow_diag}
     \end{subfigure}
     \hfill
     \begin{subfigure}[b]{0.32\textwidth}
         \centering
         \includegraphics[width=\textwidth]{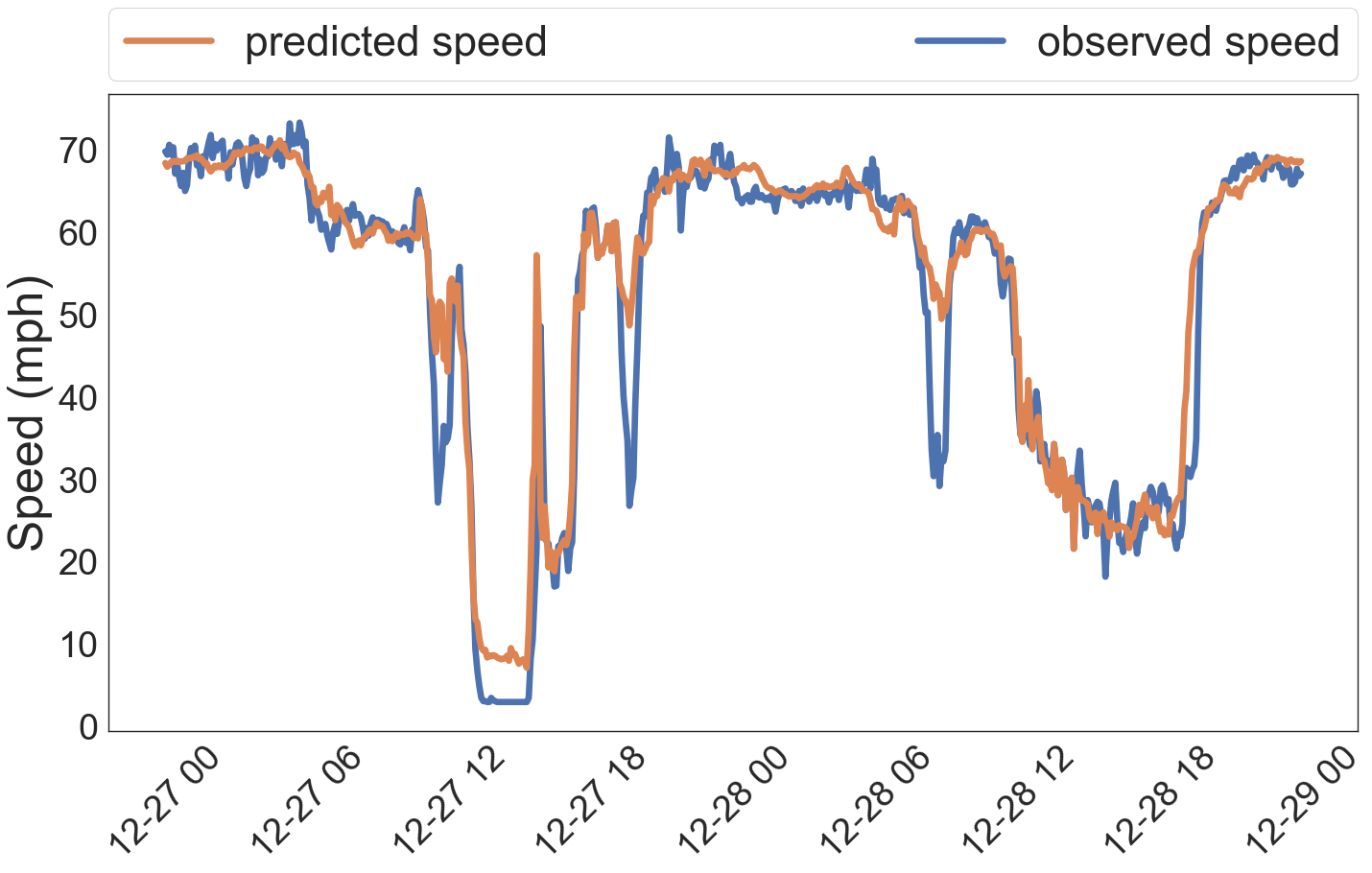}
         \caption{Speed forecasting on sensor 717322 used to estimate the traffic flow diagram of the Figure \ref{fig_flow_diag}}
         \label{fig_speed}
     \end{subfigure}
     \hfill
     \begin{subfigure}[b]{0.32\textwidth}
         \centering
         \includegraphics[width=\textwidth]{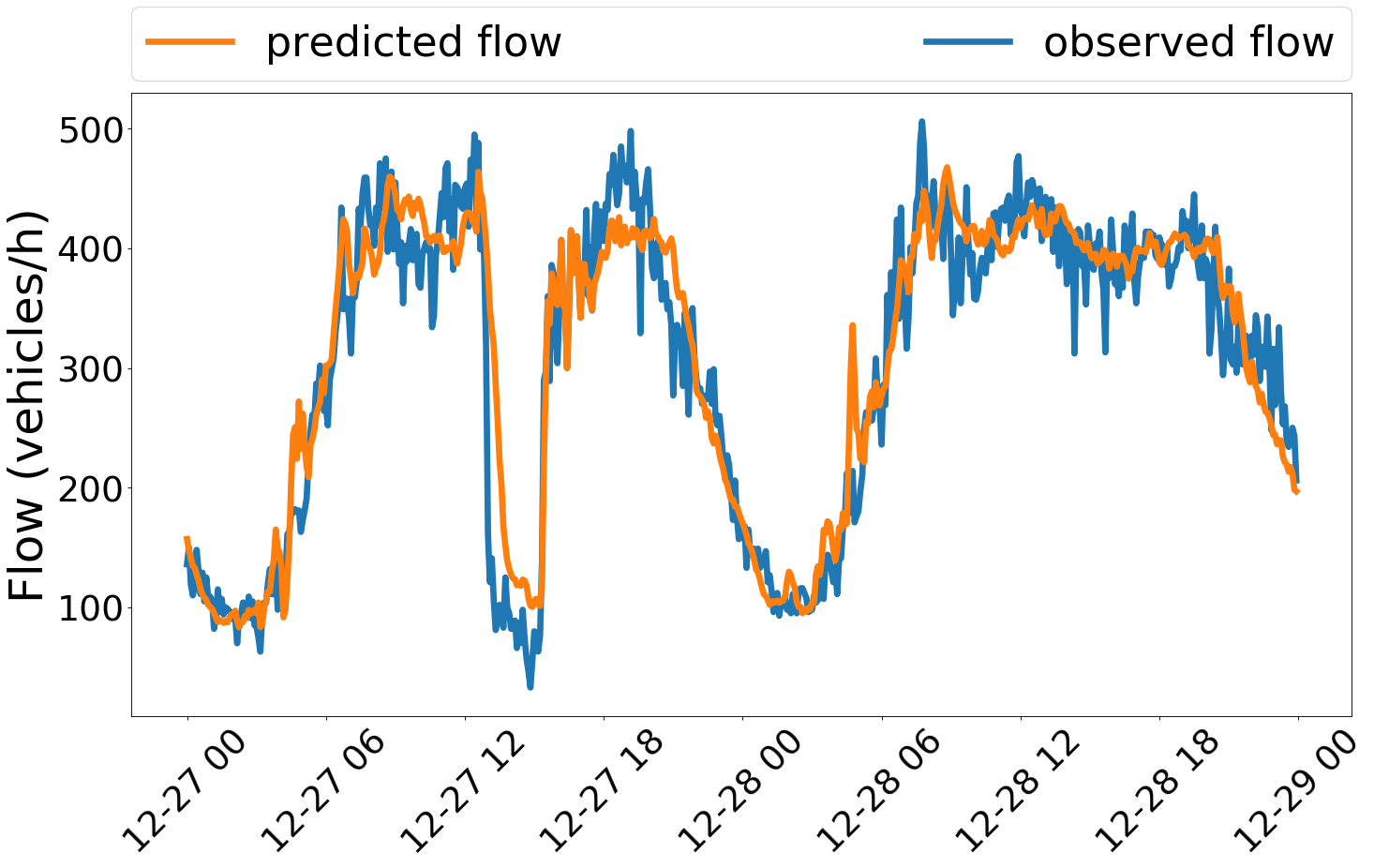}
         \caption{Flow forecasting on sensor 717322 used to estimate the traffic flow diagram of the Figure \ref{fig_flow_diag}}
         \label{fig_flow}
     \end{subfigure}
        \caption{Fundamental traffic flow diagram and corresponding speed and flow forecasting results }
        \label{fig_fundamental}
\end{figure*}

In Figure \ref{fig_flow_diag}, we show speed and flow forecasting forecasting results of a congested node (ID: 717322 located on Highway 60-E in the Los Angeles area)  
in a scatter plot. We  can see that the speed and flow forecast values closely follow the fundamental flow diagram with three distinct phases:  congestion, bounded, and free flow. This forecasting pattern of DCRNN shows that the model has learned and preserved the properties of the traffic flow. Figures \ref{fig_speed} and \ref{fig_flow} shows the time-varying speed and flow corresponding to Figure \ref{fig_flow_diag} for the same sensor 717322. 

\section{Conclusion and future work}
We developed a graph-partitioning approach to divide a large highway network into a number of partitions and trained diffusion  recurrent neural network (DCRNN) model for each of the partitions independently. We implemented an overlapping nodes improvement strategy that includes data from partitions that are geographically close to a given partition. We showed that DCRNN can be extended for multioutput learning to forecast both flow and speed simultaneously as opposed to a previous DCRNN implementation that predicted either speed or flow.
We demonstrated the effectiveness of the proposed approach using Caltrans Performance Measurement System data to model the traffic on a large  California highway network with 11,160 sensor locations. 
We analyzed the model error and showed that higher traffic dynamics caused by rapid changes in traffic behavior lead to high forecasting error. We also showed that including overlapping region can be more impactful when the partition size become small and spatially correlated nodes belong to different partitions.

The DCRNN model, once trained, can be run on traditional hardware such as CPUs for forecasting without the need for multiple GPUs and could be readily integrated into a traffic management center. Once integrated into such a center, the scale and accuracy of the forecasting techniques discussed in this work have the potential to enable more proactive decision-making as well as better decisions themselves given the capability to make large-scale and accurate forecasts regarding future traffic states.   

We plan to extend the approach to large-scale traffic forecasting with mobile device data. Our goal will be to determine whether mobile device data can act as a proxy for inductive loop data, which could  be used either to substitute for poorly working loops or to extend the scope of the monitoring to areas where loops would be prohibitively expensive. We also plan to combine DCRNN with large-scale simulations to integrate realistic speed and flow forecasts into active traffic management decision algorithms. Furthermore, we plan to develop models for route and policy scenario evaluation in adaptive traffic routing and management studies.

\section{Supporting Experiments}
A number of supporting experiments are detailed in this section.

\subsection{Missing data imputation} 
\label{sec_missing_data}


We studied three data imputation methods: (1) temporal median, which takes the median of similar time and day of the week over a period, which takes the mean of similar time and day of the week over a period of time to fill the missing value, and (3) linear interpolation. 
We evaluated these methods on a partition from the D10 Central area with 180 nodes, where 0.054\% data was missing.
We created three datasets using these imputation methods and trained the model. For training and testing data, we used $\approx$36  weeks of time series data from 1 Jan. 2018 to 13 Sept. 2018 and $\approx$10  weeks of time series data from 20 Oct. 2018 to 31 Dec. 2018. We found that the median of MAE distribution is 1.77 for the temporal mean and interpolation and 1.78  for the temporal median. The results show that missing data imputation does not have a significant impact on the forecasting accuracy. This can be attributed to the small percentage of the missing data (less than 1\%). Therefore, we replaced the missing data by temporal mean. Weekends are handled separately from normal working days.

\subsection{Graph-partitioning methods}
\label{sec_graph_partition}

\begin{figure} [t]
\centering
    \includegraphics[width=1\linewidth]{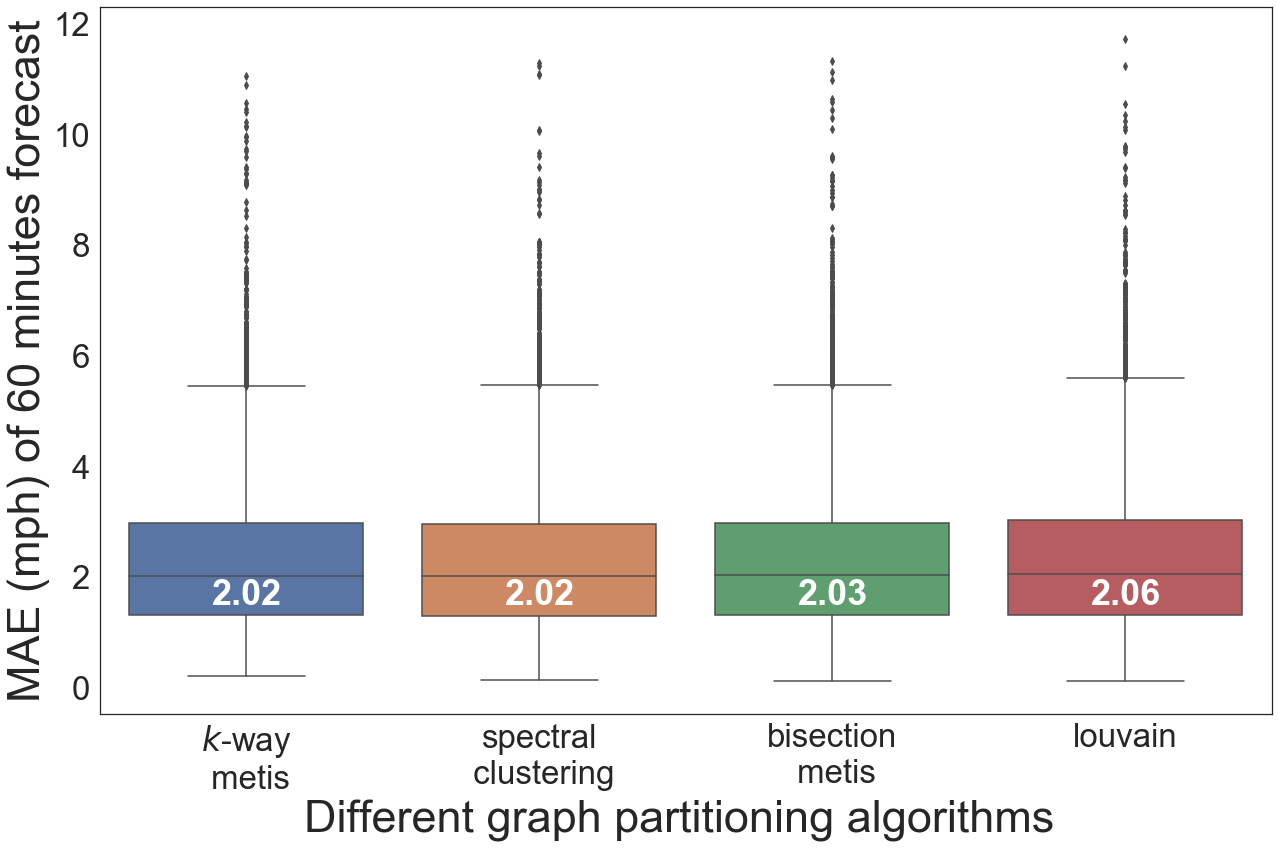}
    \caption{Impact of different graph-partitioning methods on forecasting error of the graph-partition-based DCRNN on 64 partitions}
  \label{fig_graph_part}
\end{figure}

We analyzed the impact of different graph-partitioning approaches by comparing the  default multilevel $k$-way graph partitioning approach \cite{karypis1998multilevelk} with spectral clustering \cite{ng2002spectral}, multilevel recursive bisection  \cite{karypis1998fast}, and Louvain \cite{blondel2008fast} methods. 
For the multilevel $k$-way graph-partitioning and multilevel recursive bisection approaches,  the Metis software package \cite{metis} was used; for spectral clustering, the sklearn \cite{scikit-learn} implementation was used; and for Louvain, python-louvain \cite{aynaud2018louvain} GitHub implementation was used.
Multilevel $k$-way graph partitioning, spectral clustering, and multilevel recursive bisection methods take the adjacency matrix and  number of partitions ($k$) as input and produce $k$ partitions as output. We set the number of partitions $k$ to 64 for each of the three methods. In contrast to three methods, Louvain takes the adjacency matrix as input and produces an optimized number of partitions as output, in this case 110 partitions.

Figure \ref{fig_graph_part} shows the distribution of the MAE values obtained on the test data from 4 graph-partitioning methods. The MAE values obtained with $k$-way graph partitioning (2.02), spectral clustering (2.02), and multilevel recursive bisection (2.03) are similar. The median of MAE with the Louvain method is 2.06, which is larger than that of other methods.

We adopt the $k$-way graph partitioning because of the speed advantage. In particular, the complexity of the $k$-way graph partitioning approach is $O(\varepsilon \times k)$ (where $\varepsilon$ is the number of edges in the graph and $k$ is the number of partitions), and the complexity of spectral clustering is $N^3$ (where $N$ is the number of nodes in the graph). In  a sparse graph such as ours, the number of edges is much smaller than the number of nodes. Hence, the complexity of the $k$-way graph partitioning approach is much lower than that of spectral clustering.  To perform 64 partition on a graph of 11,160 nodes, the multilevel $k$-way graph-partitioning method of Metis took only 0.030 seconds, whereas spectral clustering took 255.56 seconds.

\subsection{Comparison with other methods}
\label{sec_comparison}

We compare DCRNN to four methods: (1) lasso regression (LR) \cite{reid2016study}: $\alpha = 0.1$, the multiplier of the L1 term (implemented with sklearn Python package), (2) autoregressive integrated moving average (ARIMA) \cite{xu2017real}: order (5,1,0) is used for AR parameters, differences, and MA parameters (implemented with statsmodel python package), (3) standard feed-forward neural networks (FNN) \cite{raeesi2014traffic}:  two hidden layers and 256 neuron per layer (implemented with Keras), and (4) random forests (RF) \cite{breiman2001random}: number of trees in the forest is 100 (implemented with scikit-learn Python package).

We used the METER-LA dataset  in \cite{li2017diffusion} for model comparison.  The dataset contains 207 sensors and the time series data of 4 months collected from 1 March 2012 to 27 June  2012. As described earlier, we used 70\% (from 1 March  to 2 May  of the data for training, 10\% (from 23 May  to 4 June) of the data for validation, and  20\% (from 4 June  to 27 June) of the data for testing. Given 60 minutes of time series data, all of the models forecast for the next 60 minutes. The results showed that DCRNN outperformed all other methods. It achieved MAE of 3.60, which is lower than that of LR (7.89), ARIMA (7.73), RF (8.40), and FNN (4.49).

\section*{Acknowledgments}
This material is based in part upon work supported by the U.S. Department of Energy, Office of Science,  under contract DE-AC02-06CH11357. 
This research used resources of the Argonne 
Leadership Computing Facility, which is a DOE Office of Science User Facility under contract DE-AC02-06CH11357. 
This report and the work described were sponsored by the U.S. Department of Energy (DOE) Vehicle Technologies Office (VTO) under the Big Data Solutions for Mobility Program, an initiative of the Energy Efficient Mobility Systems (EEMS) Program. David Anderson and Prasad Gupte, the DOE Office of Energy Efficiency and Renewable Energy (EERE) program managers played important roles in establishing the project concept, advancing implementation, and providing ongoing guidance.

\section*{Author contribution}
The authors confirm contribution to the paper as follows: study conception and design: Tanwi Mallick, Prasanna Balaprakash, Eric Rask, Jane Mcfarlane; data collection: Tanwi Mallick; code implementation: Tanwi Mallick; analysis and interpretation of results: Tanwi Mallick, Prasanna Balaprakash, Eric Rask, Jane Mcfarlane; draft manuscript preparation: Tanwi Mallick, Prasanna Balaprakash. All authors reviewed the results and approved the final version of the manuscript.

\bibliographystyle{vancouver}
\bibliography{trb_template}

\end{document}